# Understanding and Visualizing the District of Columbia Capital Bikeshare System Using Data Analysis for Balancing Purposes

**Kiana Roshan Zamir (Corresponding Author)**
PhD Candidate
Department of Civil and Environmental Engineering
University of Maryland
College Park, MD 20742
Email: kianarz@umd.edu

**Ali Shafahi**
PhD Candidate
Department of Computer Science
University of Maryland
College Park, MD 20742
Email: ashafahi@umd.edu

**Ali Haghani**
Professor
Department of Civil and Environmental Engineering
University of Maryland
College Park, MD 20742
Email: haghani@umd.edu

## ABSTRACT

Bike sharing systems' popularity has consistently been rising during the past years. Managing and maintaining these emerging systems are indispensable parts of these systems. Visualizing the current operations can assist in getting a better grasp on the performance of the system. In this paper, a data mining approach is used to identify and visualize some important factors related to bike-share operations and management. To consolidate the data, we cluster stations that have a similar pickup and drop-off profiles during weekdays and weekends. We provide the temporal profile of the center of each cluster which can be used as a simple and practical approach for approximating the number of pickups and drop-offs of the stations. We also define two indices based on stations' shortages and surpluses that reflect the degree of balancing aid a station needs. These indices can help stakeholders improve the quality of the bike-share user experience in at-least two ways. It can act as a complement to balancing optimization efforts, and it can identify stations that need expansion. We mine the District of Columbia's regional bike-share data and discuss the findings of this data set. We examine the bike-share system during different quarters of the year and during both peak and non-peak hours. Findings reflect that on weekdays most of the pickups and drop-offs happen during the morning and evening peaks whereas on weekends pickups and drop-offs are spread out throughout the day. We also show that throughout the day, more than 40% of the stations are relatively self-balanced. Not worrying about these stations during ordinary days can allow the balancing efforts to focus on a fewer stations and therefore potentially improve the efficiency of the balancing optimization models.

## INTRODUCTION

The rise in popularity of sharing systems during the past years has caused different modes of sharing systems to emerge and become popular especially in urban areas. Two of the most studied transportation-related sharing systems are car-sharing and bike-sharing. Among these two, the benefits of bikes in urban settings, where the distances are short, and the parking prices are high, has caused the demand for such systems to increase. More than 1300 cities had bike-share systems in operation by July 2017 (*1*). This number is ten times more than the number reported in 2010 (*2*). In the DC, Maryland, and Virginia metropolitan (DMV) area, Capital BikeShare, operated by Alta Bicycle Share (Motivate), and mBike, operated by Zagster, are the bike-share companies in operation by July 2017.

One major factor in determining the success of any bike sharing system is the availability of bikes and docks at stations when they are needed. Based on the results of the November 2014 Capital Bikeshare Customer Use and Satisfaction Survey, 54% of Capital Bikeshare members chose availability of more docks at existing stations as the most needed Capital Bikeshare expansion option (*3*). Performance measures such as the proportion of time that a particular station is full/empty also show the need for a more efficient distribution of bikes for Capital Bikeshare. In 2015, on average, each station of Capital Bikeshare either was empty or full for 3, 8, 9, and 6 percent of the day during the first (Jan-March), second (Apr-May), third (Jul-Sept), and the fourth (Oct-Dec) quarter, respectively (*4*). This is mainly due to an imbalanced system. Many reasons might cause the system to become imbalanced. Some of which are (*5*):

- Flow patterns,
- Availability and frequency of other transportation modes,
- Altitude of stations,
- Weather, and
- Traffic conditions.

Balancing has been addressed in the literature by two approaches. The first one is by introducing a pricing mechanism that imposes rewards to users who pick up their bikes from stations that have excess bikes or drop them off at stations that have shortages. The second approach is by repositioning the bikes in the system. This involves the movement of bikes across the networks for maintaining an even distribution (*6*).

Solving the balancing problem for mid/large-size instances can be easier if a general understanding of the system is available since operation research models are sensitive to the number of variables. Having knowledge of the system can help reduce the number of variables and consequently the problem size. Many of the pace-improving solution methods and heuristics that are tailored to a problem require insight of the system. For example, one may effectively shrink the search space for a model by adding valid constraints such as using larger trucks for balancing stations which tend to become more imbalanced, or removing stations that are relatively self-balanced during the balancing period. Such self-balanced stations need relatively few rebalancing which can be done during non-peak hours such as night. Data mining techniques can provide such information for operation research models.

Data visualization, which is defined as the effort of placing data in a visual context, can assist in better understanding the problem parameters/data. When we have many data points, visualization of the data becomes challenging, and data consolidation is required. Data consolidation can be done by clustering. Clustering is a well-known technique for identifying some inherent structure in an unlabeled data. The objective of clustering is to group the data set into homogenous groups in which objects within a group have similar features and dissimilarity of the objects among different groups is maximized (*7-8*).

The purpose of this study is to use data mining techniques such as clustering for grouping the bike-sharing stations that have similar temporal activity patterns regarding pickups and drop-offs along with visualization. The result of this grouping can be used as a tool for analyzing the balancing efforts needed, and improving the design of stations. The enhancement can be achieved by providing managerial insights such as expansion recommendations for areas where the existing stations have higher demands. This study also helps in gaining a better understanding of the urban mobility of District of Columbia's residents. To the best of the authors' knowledge, no research exists that has attempted to visualize the pickup and drop-off profiles of the Capital Bikeshare stations using this methodology. This paper also identifies the stations that are imbalanced and those that, in expectation, require little or no balancing within a period based on the Average Daily Maximum Shortage (ADMS) or Excess (ADME) of bikes. Throughout the paper, we will give some useful managerial insights that can be drawn from the clustered and categorized stations.

The structure of the paper is as follows. First, we provide a brief literature review on studies related to bike-share systems. The data set used, and the clustering and categorization methods are described in the next section. The third section is dedicated to the results. Finally, in the last section, we provide the summary and conclusions.

## LITERATURE REVIEW

The concept of the bike-share system was introduced in 1965 in Amsterdam (*9*) and has evolved by the introduction and development of many technologies such as smartphones. Since the introduction of bike-share systems, much research has been dedicated to different aspects of these systems. Some of these studies are descriptive and mine the data to get a better understanding of the operations. While others are prescriptive and suggest ways to improve the operations. The most common prescriptive model is related to balancing. Various studies have focused on mathematical optimization and have addressed the rebalancing problem by developing mathematical models for repositioning the bikes in the system *(6,10-16)*.

A majority of the descriptive models develop prediction models for the departure and arrival rates, and the bike availabilities. Froehlich et al. (*17*) provided a spatiotemporal analysis of bicycle usage and several predictive models for the availability of bicycles for Barcelona's bike-share system, "Bicing." Faghih-Imani et al. (*18*) identified the determining factors of bike-share usage for Montreal's bike-share system, "BIXI", by developing models for prediction of the arrival and departure rates of stations. Yoon et al. (*19*) developed a spatiotemporal prediction model for Dublin's bike-share system which can be used to help users of bike-share systems in planning their trips. Chen et al. (*20*) proposed a class of algorithms which use the Generalized Additive Models for predicting bike availability. Giot and Cherrier (*21*) examined various regressors for predicting the usage of bike-share systems. Li et al. (*22*) proposed a hierarchical prediction model for predicting the rented and returned bikes from and to each station.

Some other descriptive research investigates the effects of changes to different impacting factors of the bike-share systems. Lathia et al. (*23*) examined the impact of the change in the user-access policy of London's bike-share system, "Barclays Cycle Hire", which was modified to allow for casual usage. Corcoran et al. (*24*) investigated the effects of weather and calendar events on the spatiotemporal dynamics of Brisbane's bike-share system.

Some other descriptive research has focused on analyzing the data gathered from these systems. Owen et al. (*25*) evaluated Washington D.C.'s Capital Bikeshare and New York City's Citi Bike crowdsourcing effort. Vogel et al. (*26*) used the ride data from Vienna's bike-share system, "Citybike Wien", to confirm the hypothesis that activity patterns and station's location are correlated using clustering. Bargar et al. (*27*) developed a visual analytical tool for comparing the usage patterns of different bike-share systems and developed a clustering algorithm for identifying similar trips. Randriamanamihaga et al. (*28*) used Poisson mixture models and expected maximization to cluster origin/destination flow data for the Paris bike-share system, "Vélib". O'Brien et al. (*29*) used data from bike-share systems around the world and classified these systems based on their aggregated, spatial, and temporal characteristics. Etienne and Oukhellou (*30*) used model-based count series to cluster "Vélib's" stations based on their usage.

Particularly, for the Capital Bikeshare, there has been some descriptive research that tries to visualize and analyze the data from Capital Bikeshare. Chen et al. (*31*) developed a method for inferring spatiotemporal bike trip patterns using the stations' pick up and drop offs data. Coffey and Pozdnoukhov (*32*) explored the relationships between the temporal usage of bike-shares with Twitter's tweets and Foursquare check-ins. Ahmed (*33*) clustered the stations based on the number of rentals and labeled the stations based on their usage into commuter, leisure, residential, and tourist or mix. Chen et al. (*34*) proposed a dynamic clustering scheme that groups nearby stations based on their usage pattern. This clustering along with the Monte-Carlo simulation is used to predict the over-demand probability of each cluster.

In this study, we focus on clustering the Capital Bikeshare stations based on their temporal profiles. We also classify the stations based on the amount of attention they would need during the balancing process using two indices that measure how much each station is self-balanced on expectation.

## DATA AND METHODOLOGY

### *Data*

Capital Bikeshare is a program operated by Motivate International, Inc., and is jointly owned and sponsored by the District of Columbia, Arlington County, the city of Alexandria, VA, and Montgomery County, MD. It offers different membership packages, which vary based on membership period. The first 30 minutes of each trip is free for members, and additional surplus is charged for every 30 minutes afterward.

Capital Bikeshare was the largest bike-share system in North America until May 2013. It currently is the fourth largest in North America. As of July 2016, Capital Bikeshare had 3050 bikes and 391 stations. In the presence of this many stations, analyzing and visualizing each station by itself is impractical. To better visualize the Bikeshare data, the stations should be aggregated. One approach to aggregate the stations is clustering based on similarity measures. 2015 trip history data for Capital Bikeshare is used in this study. The data is divided into four quarters (The first quarter being Jan-March) to capture the seasonal effects of pickup and return profiles, and clustering is performed separately for each quarter using the profile as the similarity measure. The data set includes duration, start date, end date, start station, end station, bike#, and member type for each rental (*35*). Stations' locations are obtained from the District of Columbia open data (*36*).

Table 1 provides summary statistics for each quarter of this data set. Figure 1 provides the average number of pickups and drop-offs for each hour along the week for each of the quarters. It can be seen that during weekdays, three peaks exist: morning, noon, and afternoon; the afternoon peak is the greatest. During weekends, we do not see significant peaks. The average profile is smoother in comparison to weekdays. The peaks and drops in different quarters' profiles happen more-or-less during the same hours of the day. However, the magnitude of the average usage is different due to the different total number of trips during different quarters as evident in Table 1 (presumably due to weather conditions). The hourly drop-off profile is similar to the hourly pickup profile but with a small shift. This is expected since the average trip durations are very short according to Table 1.

**TABLE 1 Summary statistics for each quarter of 2015 Capital Bikeshare data**

|  | **First quarter** | **Second quarter** | **Third quarter** | **Fourth quarter** |
|---|---|---|---|---|
| **Number of trips** | 431,465 | 999,073 | 1,048,576 | 706,004 |
| **Number of stations** | 347 | 353 | 355 | 356 |
| **Average duration of trips (mins)** | 13 | 19 | 19 | 16 |

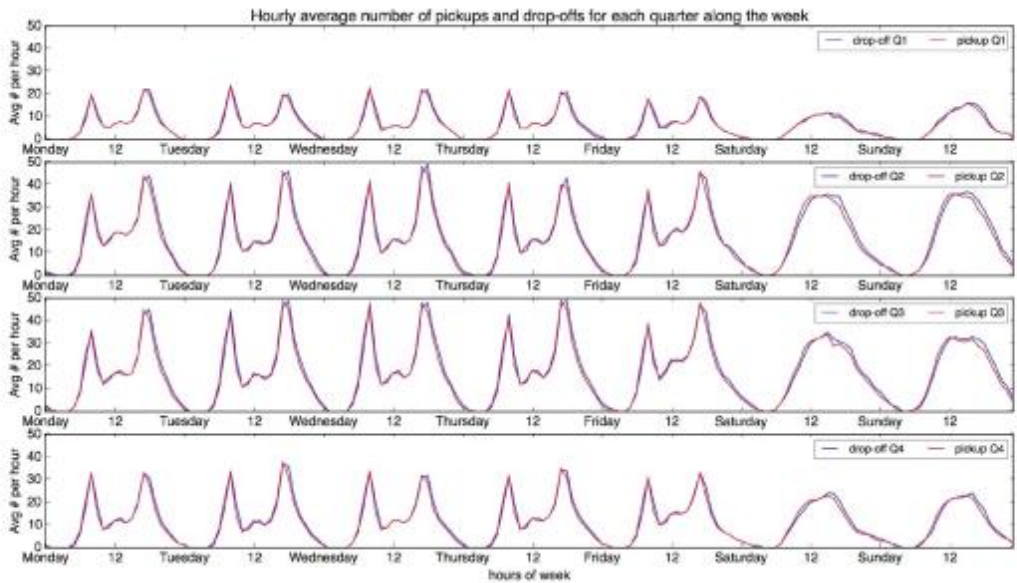

**FIGURE 1 Hourly average of pickups and drop-offs along different hours of the week per quarter**

*Methodology*

In this paper, two different visualizations and clustering of the bike-share data is provided. One of them is the hourly pickup/drop-off trends (profiles) of stations. The other is based on the imbalance severity of each station. The stations are clustered together based on their general profile patterns to reach the first objective of showing temporal pickup/drop-off trends for the stations. For the second objective, the stations are categorized based on their ADMS and ADME for the day. ADMS and ADME are defined in detail later in this methodology section.

*Clustering Methodology for Visualizing Daily Temporal Trends of Stations*

The clustering is done based on the temporal (hourly) profile of the percentage of pickups/drop-offs during different hours of the day. We chose the percentage of pickups/drop-offs instead of the number of pickups/drop-offs because different stations have different capacities. This clustering intends to group the stations that have similar trends together and not those with similar sizes. The clustering methodology has five main steps: 1) data cleaning; 2) aggregation of data; 3) normalization of the stations' pickups and drop-offs; 4) removing stations with non-typical behaviors; 5) identification of an appropriate number of clusters in K-means. Each step is described in more detail in the following.

Data cleaning is an intrinsic part of any data analysis. Here, data cleaning refers to both the removal of data errors caused by imperfect data collection and removal of anomalous data points (*37*). The latter is particularly important when the purpose of data analysis is clustering. The presence of "noises" can distort the result of clustering abruptly (*38-39*). Several methods are discussed in (*37*) for identification and removal of noises. Here, a similar methodology described in Portnoy et al. (*40*) is used to find stations with anomalous behaviors. For clustering, in each quarter, trips and stations were removed from the data set if they had the following attributes:

- Trips with less than 60 seconds duration
- Trips with duration less than 120 seconds that start and end at the same station
- Stations with negligible pickups and returns: Stations with daily averages of less than five pickups or drop-offs are removed from pickup data set and drop-off data sets, respectively.
- Stations that show non-typical behavior. These are identified by the size of their cluster after applying the K-means and Agglomerative hierarchical clustering algorithms. We remove outlier stations that are by there own and far from the main clusters as described in Portnoy et al. *(40)*. The removed stations should be analyzed separately.

After cleaning the data, we aggregate the data on an hourly basis. For each day of the quarter and each station, we aggregate the pickups/drop-offs on an hourly basis by assigning all the pickups/drop-offs between two hours to the next integer hour. We then normalize the total hourly pickups/drop-offs by dividing them by the total number of pickups/drop-offs during that day. After this step, we identify and remove the stations with non-typical behaviors using the K-means and Agglomerative hierarchical clustering algorithms as mentioned before.

Euclidean distance between the aggregated normalized hourly pickups/drop-offs is selected as the dissimilarity measure for clustering. We perform the main clustering using the K-means algorithm developed by Hartigan and Wong (*41*). K-means produced better results compared to hierarchical clustering on our data set.

The last step is validating the result of clustering analysis and finding the suitable number of clusters (K). In most cases, a combination of compactness and separation measures are used for evaluating the clusters (*38*). Compactness measures the homogeneity within the clusters and separation measures the degree of separation between clusters. Davies-Bouldin index, Silhouette Width, and Dunn Index are some of the well-known internal validation measurements in the literature that are used in this study for evaluating the quality of clusters and finding the suitable number of clusters.

*Categorizing the Stations based on Maximum Shortage (ADMS) or Excess (ADME)*

ADMS measures the maximum shortage of a station during a given period averaged over all days of a quarter. ADME measures the maximum excess. To calculate the maximum shortage of a station between hours $t_1$ and $t_2$, we sort all of the pickup and drop-offs for the station that happen between $t_1$ and $t_2$ based on their occurrence time so that the soonest pickup or drop-off is first. We start with the shortage being 0 at $t_1$. Then, we process each drop-off and pickup one at a time. Every time we see a pickup, we increment the shortage, and if we see a drop-off we decrement the shortage. Once the entire list of pickups and drop-offs are processed, we take the maximum positive number we had for the shortage as maximum shortage for that period of that particular day. The smallest negative number is the maximum excess. Figure 2 illustrates how these values are found. After calculating these two measures for all of the days in a quarter, the daily average of the shortage would be ADMS and the daily average of the excess would be ADME.

These values can act as proxies for the amount a station is self-balanced during the given period. In the case that exactly after each pickup, a drop-off happens, the maximum shortage would be 1 and the maximum excess would be 0 (Day $i$ in Figure 2). In another case, if we have $n$ pickups and $n$ drop-offs but first all of the pickups happen then all of the drop-offs, the maximum shortage would be $n$ while the maximum excess would be 0. Therefore, it is wise to look at ADMS and ADME together and also individually. Note that this measure is greatly affected by the duration of the period that is selected for studying. If both ADMS and ADME are small for a duration of 4 hours, we then have a relatively self-balanced station during those 4 hours.

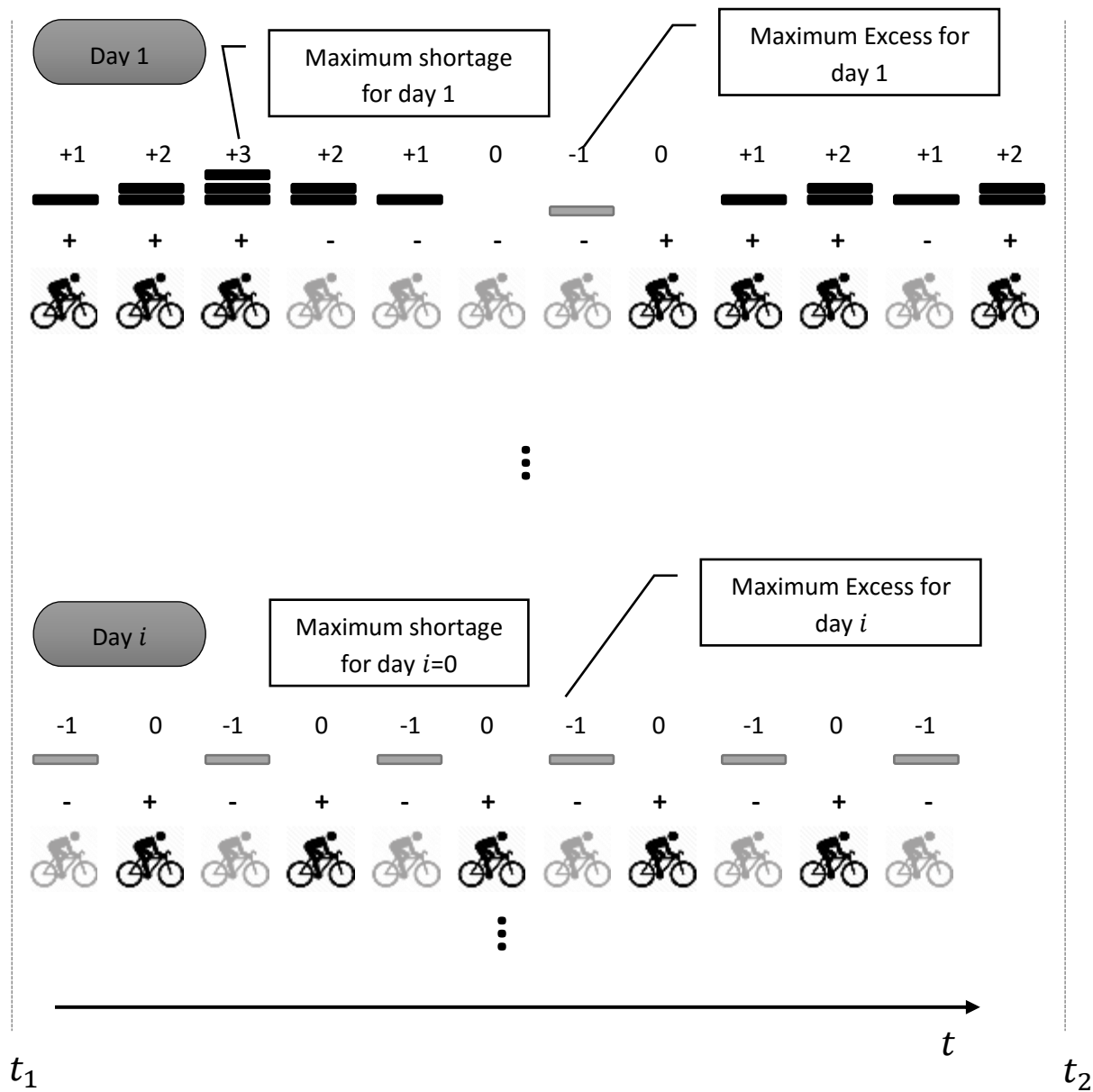

**FIGURE 2 Illustration of how the maximum shortage and excess are calculated for different days. The average of these is then reported as ADMS and ADME.**

## RESULTS

The clustered stations and the profiles for the center of clusters for each quarter are calculated and graphed. Due to space limitations, we only include the results for July-September (quarter 3) that is the quarter with the most trips. However, the high-resolution profiles for all quarters can be found in (*42*). Also, the clustered maps for quarter 1 can be found in (*43-44*). The maps for quarter 2 can be found in (*45-46*). The third quarter's maps are available in (*47-48*). Finally, the maps for the fourth quarter are available in (*49-50*).

Figure 3 shows the result of clustering for (a) drop-off and (b) pickup of the third quarter (July-September). Stations that are represented with the same color are in the same cluster with relatively similar temporal patterns. We can see that stations that are geographically close to each other have a similar pickup and drop-off profiles which confirms the past studies (*26*).

Figure 4, illustrates the various clusters' centers for drop-off weekday (a) and weekend (b), and pickup weekday (c) and weekend (d), respectively. Due to significant variability of pickups and drop-offs during weekdays and weekends, their profiles are shown in separate graphs. Each graph is drawn based on 24 numbers that indicate the cumulative color percentage of drop-offs or pickups for each one-hour interval. For instance, around 20 percent of cluster 7 (light blue stations)'s drop-offs happen between 6-7 p.m. during weekdays (Figure 4-a). We can see that pickups and drop-offs are more evenly distributed during weekends (Figure 4-b and 4-d), whereas in weekdays pickups and drop-offs are concentrated around three points: morning, lunch, and afternoon peaks (Figure 4-a and 4-c). This could indicate less balancing efforts are needed for weekends

Stations are grouped into 8 clusters concerning their drop-off behavior (Figure 4-a). The largest cluster (cluster 4) has 70 stations. Around 80% of the stations belong to three clusters (cluster 4: yellow, cluster 1: brown, and cluster 5: green in Figure 4-a&b). Cluster 3 (pink) stations are mostly located in downtown D.C in which the land-use is mostly commercial according to the DC Office of Zoning (DCOZ). As expected, these stations have many drop-offs in the weekday mornings. Presumably, these are commuters who use bikes to get to their work in the morning. Around 40 percent of drop-offs of the stations in this cluster happen between 7 to 10 a.m. Morning peak happens between 8 to 9 a.m. for most of the clusters. Brown stations (cluster 1) are stations that are far from the downtown area and are located in residential areas. These stations have many drop-offs in the afternoon. Usually, afternoon peak happens between 5 to 8 p.m. which indicates that afternoon peaks are more spread out compared to morning peaks. Yellow stations (cluster 4) are located between pink and brown stations. They are not very close to downtown nor far from it (located in areas with a mixture of commercial and residential land use). A high percentage of drop-offs can be seen for these stations during both morning and afternoon peaks. Figure 3-a and Figure 4-a give a big picture of the balancing efforts needed. For instance, bikes may need to be removed from stations that are shown with pink color due to the high percentage of drop-offs in the morning and moved to stations with brown color. This classification can be incorporated into dynamic repositioning models by adding information retrieved from these figures as constraints to the model, for instance, banning the optimization model from removing bikes from the stations that will need bikes in the next interval to balance the neighborhood stations. Balancing efforts are further explored and explained in the maximum surplus and shortage analysis.

Based on Figure 3-b and Figure 4-c&d stations are grouped into 7 clusters concerning their pickup behavior. The majority (83%) of the stations belong to four clusters (cluster 4: yellow, 5: pink, 3: brown, and 2: blue in Figure 4-c&d). Pink stations are mostly located in downtown D.C. and have high percentages of pickups during the afternoon. Similar to drop-off, morning peak happens between 8-9 a.m. for most of the clusters and afternoon peaks are distributed in a larger interval compared to morning peaks. Brown stations (cluster 3) are stations that are located between downtown stations and stations distant from the downtown. These stations have a similar percentage of pickups and drop-offs in the morning and afternoon peak. Blue stations (cluster 2), are stations located far from downtown. They have higher percentages of pickups in the morning.

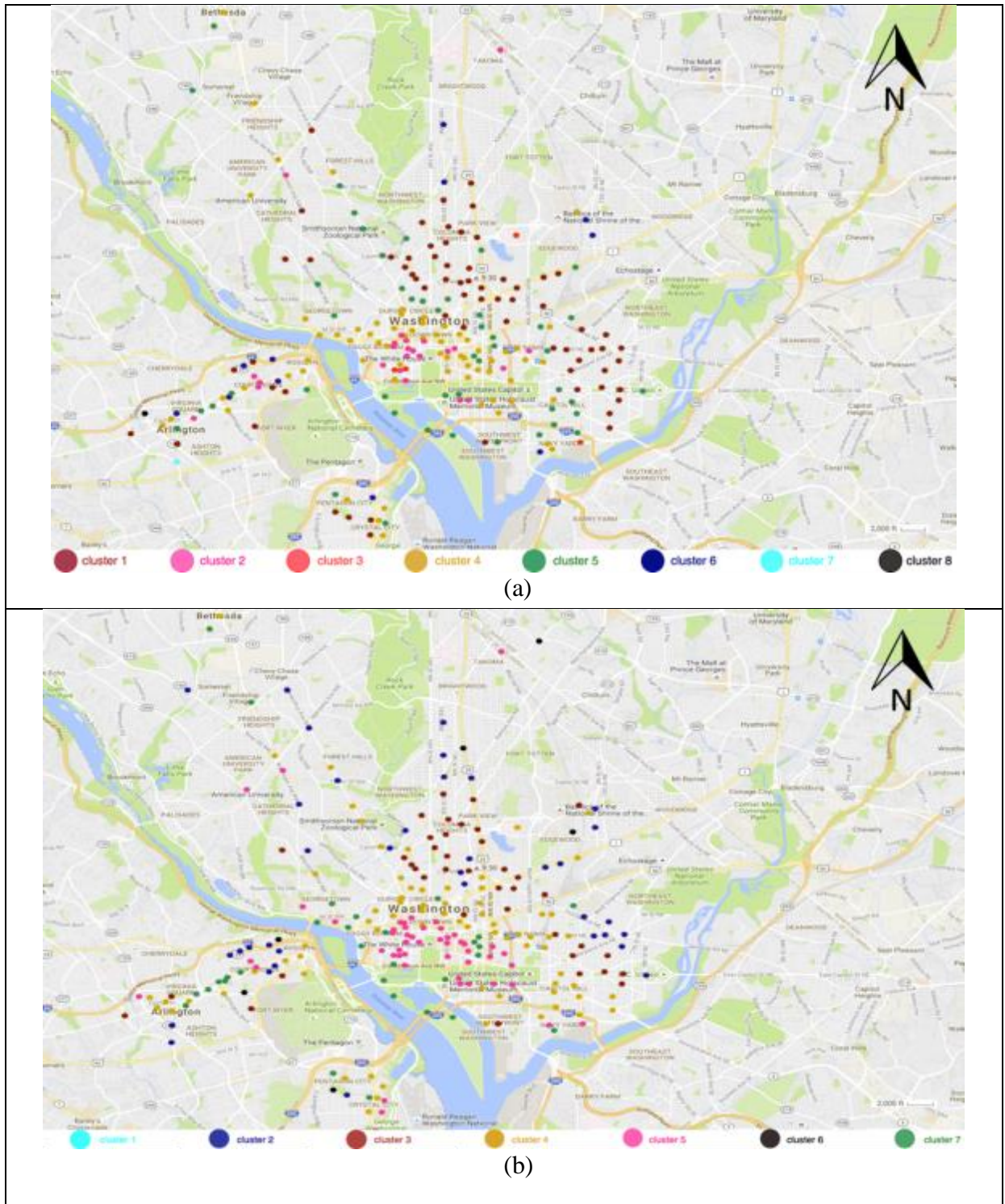

(a)

(b)

**FIGURE 3 3rd quarter (Q3) drop-off (a)** *(47)* **and pickup (b)** *(48)* **clustered stations**

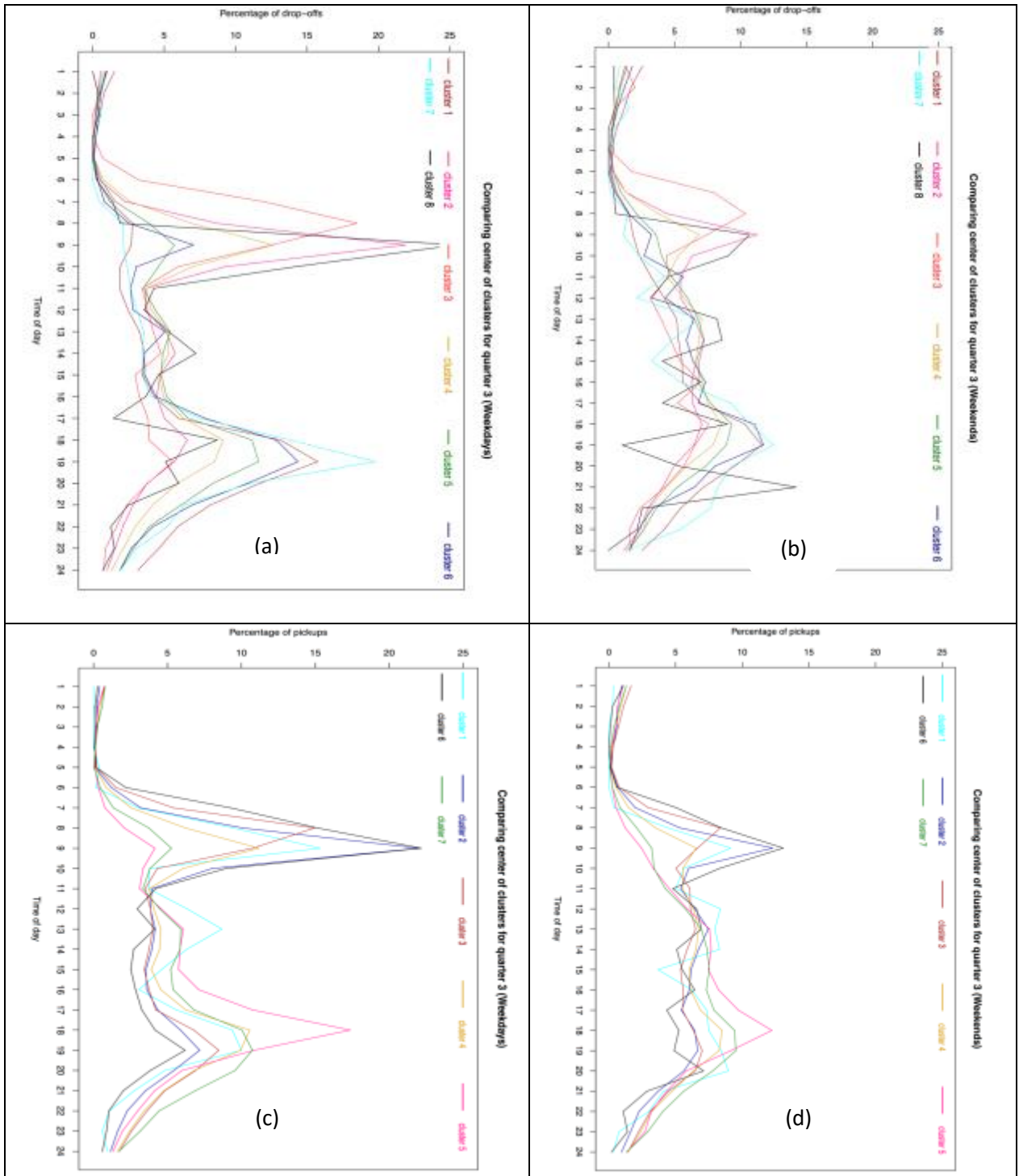

**FIGURE 4 Drop-off trend for clustered stations during weekdays (a) and weekends (b) for; and the pickup trends for clustered stations during weekdays (c) and weekends (d) of Q3.**

The categorization results based on ADMS and ADME for identification of balanced stations are done for three different time periods during the third quarter to illustrate the effect of time-of-day on balancing requirements. Each station, based on its ADMS or/and ADME value, is assigned to one of eight categories each reflecting a different degree of imbalanced stations. The first category reflects highly balanced stations with ADMS or/and ADME less than or equal to five. Parts (a), (b), and (c) of Figure 5-Figure 7 illustrate the categories for the stations when (a): only maximum excess (ADME) is used for categorization; (b) only maximum shortage (ADMS) is used for categorization, and (c) when both maximum excess (ADME) and shortage (ADMS) are used for categorization.

By looking at the results for only ADME (a) and only ADMS (b), we can improve the system by transferring bikes from the stations with high ADME to those with high ADMS. By looking at part (c) (Both ADME and ADMS together), we can see which stations are already self-balanced (colored cyan) since, during that time period, both ADME and ADMS are both below 5. This means on average, during that time period, the pickups and drop-offs are balancing each other as the maximum excess/shortage from the beginning of the time period is very small. In other words, if the station has five bikes and five empty racks, it is likely that it would not get full or empty. We have colored the stations that fall within this category using cyan that makes them almost invisible meaning that very little focus should be paid to these for balancing purposes.

The first period studied is the entire 24hrs of the day (Figure 5). ADMS and ADME for 24hrs is a relatively robust measure that can be used as a proxy for finding the maximum size for stations' capacities. If the size of a station is equal to the ADMS + ADME, we expect to need no balancing at all during the day. It is interesting to see that about 41% of the stations are self-balanced. For example, we can see that many stations surrounding Arlington require no more than ten docks (Figure 5). The category of the station could also be used in assisting users in picking a nearby station. For example, the maximum average daily shortage and excess happens at the station located at the intersection of 13th st. & H St. based on Figure 5-c. However, after examining Figure 5-a&b, we can see that the main reason for this is due to having the maximum ADME and not due to a large ADMS. Incentivizing users to pick up more bikes from, or drop-off fewer bikes at this station can help reduce this peak. Adding new station near this location is another solution for reducing the incoming flow to this station.

Figure 6 is the visualization for the stations' categorizations for the morning peak-hour period (i.e. 6:00 a.m. – 10:59 a.m.) and Figure 7 is for the afternoon peak-hour period (i.e. 4:00 p.m. – 7:59 p.m.). Since these two periods are subsets of the first period (entire 24hrs of the day), the maximum shortage and excess during these periods for each station are less than or equal to the same measure during the 24hr period. During morning peak hours, about 53% of the stations are self-balanced. However, most of the maximum daily shortages and excesses happen during the morning-peak period as the categories for the stations during the morning peak are quite similar to those from the 24hr period. This indicates that although on average we observe more trips during the afternoon (Figure 1), these trips are more balanced compared to trips during the morning peak. This is also in line with our results in the previous section that morning peak happens in a shorter interval compared to afternoon peaks. It also indicates that more rebalancing efforts are needed during the morning peaks.

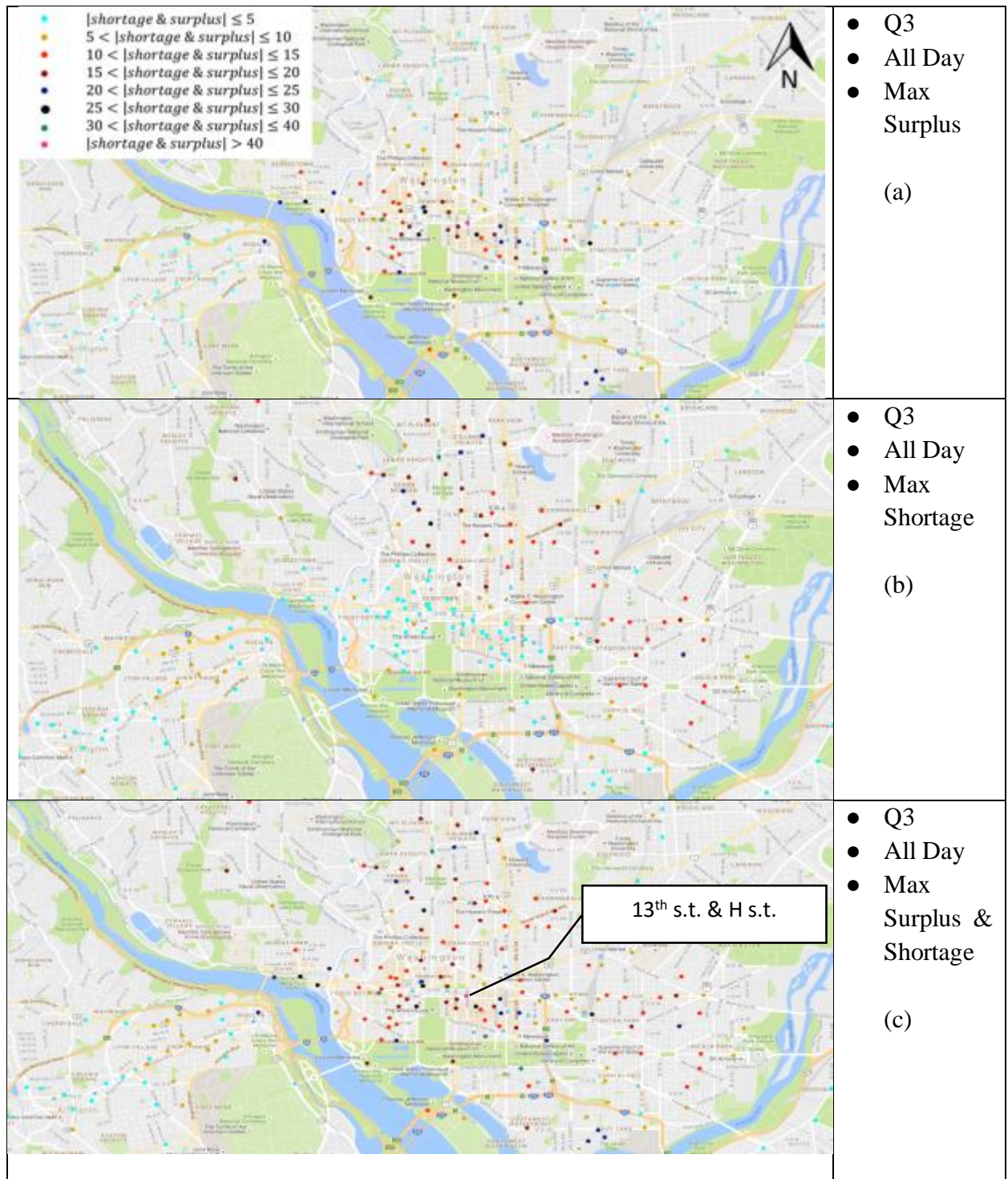

**FIGURE 5 Categorization of stations based on daily average maximum shortage and surplus during 24hrs** *(51-53)*

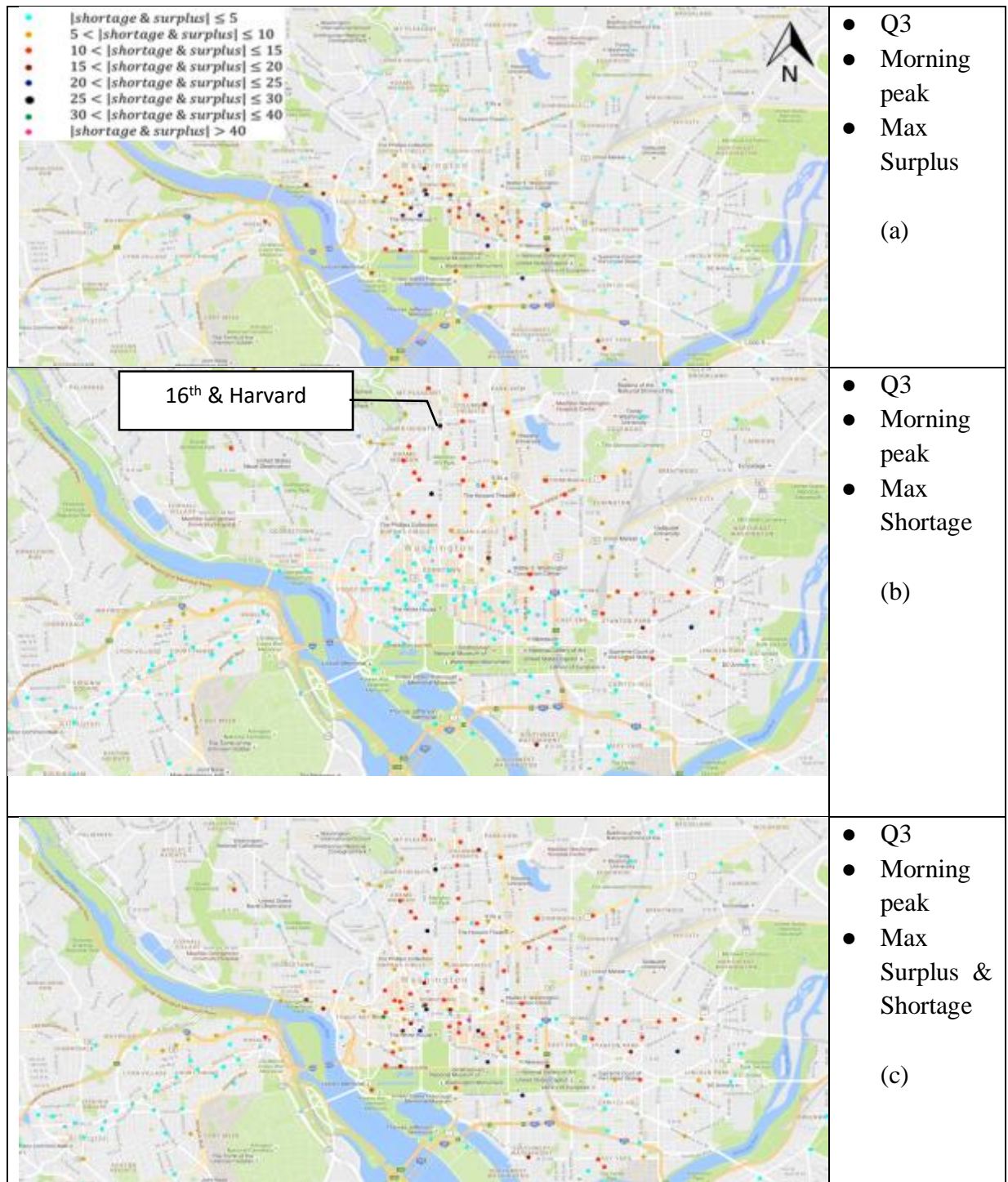

**FIGURE 6 Categorization of stations based on daily average maximum shortage and surplus during morning peak hours (6:00-10:59 a.m.) (*54-56*)**

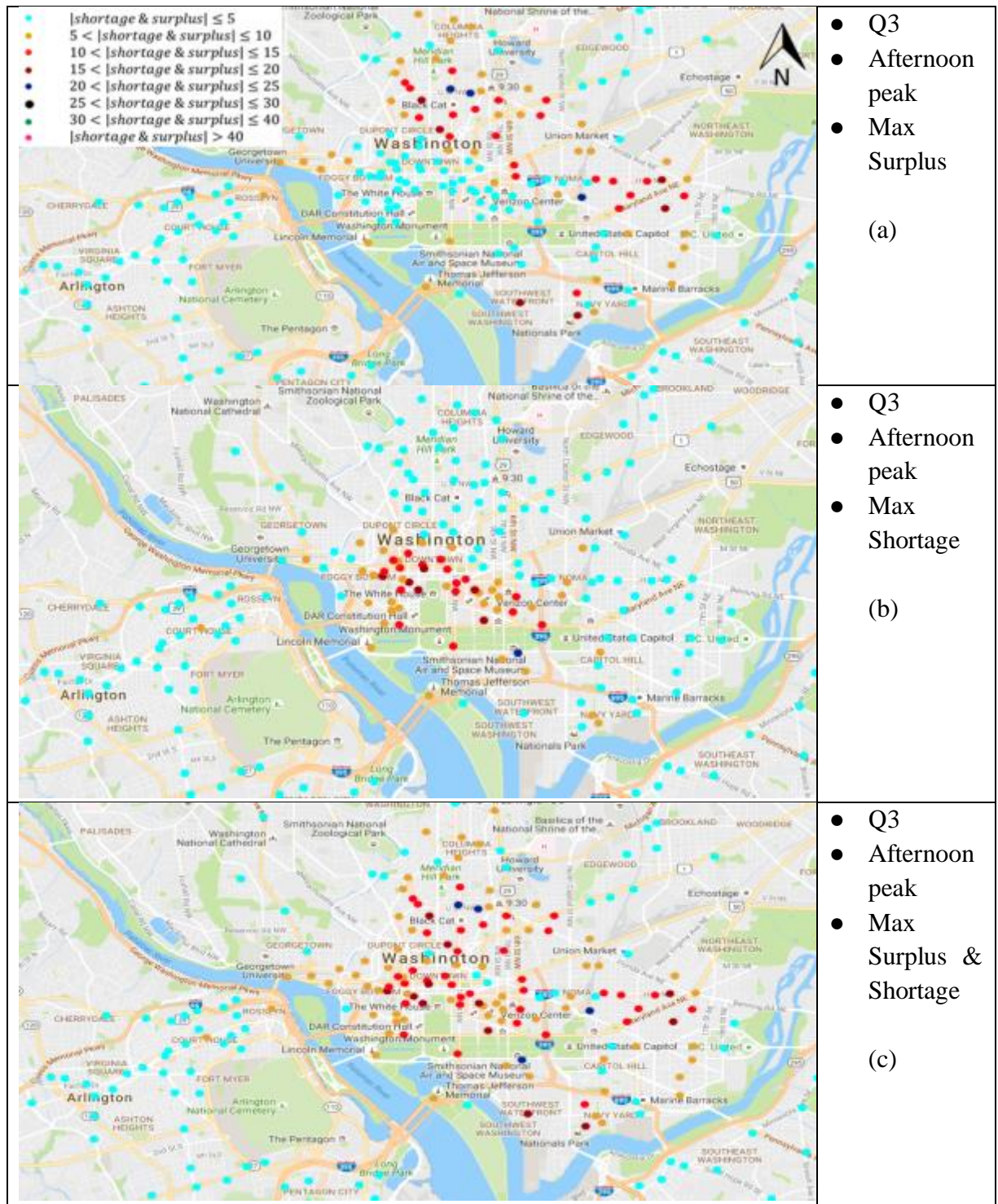

**FIGURE 7 Categorization of stations based on daily average maximum shortage and surplus during afternoon peak hours (4:00-7:59 p.m.)** *(57-59)*

For shorter time periods, we can suggest operational policies for balancing. For example, to address the maximum shortage at the intersection of Harvard St. and 16th St. (Figure 6-b) that is categorized as black (between 20-25), we can assign a truck to pick up bikes from surplus stations that are close to it until 20-25 bikes are picked up. Since none of the close stations surrounding this station have significant surpluses, the truck has to visit multiple stations (at least 5).

Also, by looking at figures 6 and 7, stations that have a high average surplus with neighboring stations with a relatively low surplus can be identified. Through the identification of these stations, authorities can assist in assigning ideal drop off stations to bikers so that the likelihood of not having any vacant bike racks/docks is minimized. Authorities may also propose incentives such as extra free minutes for users that drop off their bikes in the stations with shortages.

Figure 8 illustrates the categorizations of the stations during non-peak middle-of-the-day hours (12:00 p.m. – 3:59 p.m.). During these hours, most of the stations (about 87%) are almost self-balanced, and a relatively small balancing crew should be able to perform the required balancing.

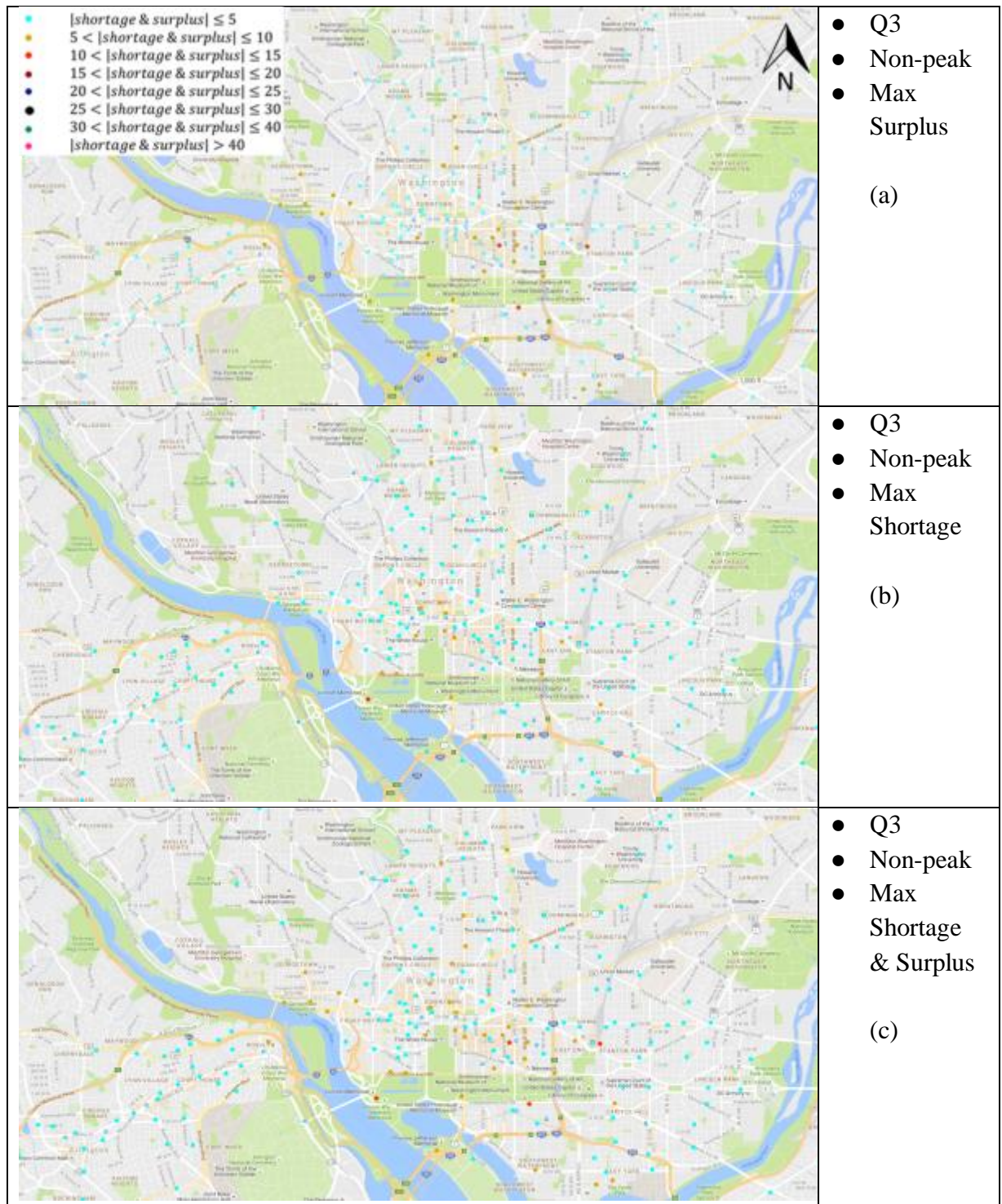

**FIGURE 8 Categorization of stations based on daily average maximum shortage and surplus during non-peak hours (11:00 a.m. -15:59 p.m.) (*60-62*)**

## CONCLUSIONS

As part of ongoing efforts to improve bike-share systems' efficiency, this study uses clustering methods for consolidation of the data retrieved from Capital Bikeshare that is among the top largest bike-share systems in North America. The methods employed can assist in visualizing the general trends in pickup and drop-off. It can also assist in identifying balancing requirements needed during different times of the day. Stations are grouped based on their imbalance severity during different periods of the day. This approach can be used for identifying stations with high chances of getting full or empty and subsequently coping with these problematic stations using different techniques such as applying pickup/drop-off incentives from/to these stations.

Findings show that during weekdays most of the pickups and drop-offs happen during the morning-peak and afternoon-peak for Capital Bikeshare whereas during weekends the pickups and drop-offs are more uniformly distributed. Also, on average, many stations (about 41%) are relatively self-balanced during the day (i.e. the difference between their pickups and drop-offs is negligible). These stations may only need to be balanced during off-peak or idle periods such as night.

Moreover, this study brings new insights that can be useful in repositioning optimization models. Most balancing optimization models assume that trucks do simultaneous pickups and drop-offs. However, as illustrated in this paper, due to the similar trends of nearby stations, simultaneous pickup and drop-off could cause longer routes as the distances between stations with deficiencies and stations with surpluses are relatively large. An alternative strategy for balancing may be to first solely perform pickups until the truck's capacity is met and then perform the drop-offs.

This study opens venues for much future research related to data analysis and operations research. One can use the insights provided in this paper for developing different balancing policies, various promotion mechanisms for facilitating self-balancing, and developing bike-share usage prediction models for the Capital Bikeshare. Also, operational research models can be proposed that take advantage of demand clusters and evaluate the alternative strategy of performing pickups and drop-offs separately. This approach can shrink the solution space for the rebalancing greatly. However, the quality of the solutions must be tested for different networks.